\documentclass[conference]{IEEEtran}
\IEEEoverridecommandlockouts
\usepackage{cite}
\usepackage{amsmath,amssymb,amsfonts}
\usepackage{algorithmic}
\usepackage{graphicx}
\usepackage{textcomp}
\usepackage{xcolor}

\usepackage[caption=false,font=footnotesize]{subfig}

\usepackage{amsmath, amssymb, amsfonts}
\usepackage{algorithm, algorithmic}

\usepackage{cite}
\usepackage{enumitem, textcomp, soul, xcolor}
\usepackage{multirow, colortbl}
\usepackage{epigraph}
\usepackage{hyperref}
\usepackage{tabularx, float, longtable}
\usepackage{booktabs, cleveref, threeparttable}

\usepackage{microtype}

\def\BibTeX{{\rm B\kern-.05em{\sc i\kern-.025em b}\kern-.08em
    T\kern-.1667em\lower.7ex\hbox{E}\kern-.125emX}}
\begin{document}

\title{A Dual-Cam Parallel Elastic Actuator with Shared Gas-Spring Compensation for Humanoid Ankles\\
\thanks{This work was supported by European Union’s Horizon Programmes under Grant 101070596 (euROBIN). (\textit{Corresponding author: Jingcheng Jiang.})\par\vspace{2pt}\textcopyright~2026 IEEE. Personal use of this material is permitted. Permission from IEEE must be obtained for all other uses, in any current or future media, including reprinting/republishing this material for advertising or promotional purposes, creating new collective works, for resale or redistribution to servers or lists, or reuse of any copyrighted component of this work in other works.}
}

\author{
\IEEEauthorblockN{Jingcheng Jiang}
\IEEEauthorblockA{\textit{Humanoids and Human Centered}\\
\textit{Mechatronics (HHCM) Research Line} \\
\textit{Istituto Italiano di Tecnologia}\\
Genova, Italy \\
jingcheng.jiang@iit.it}
\and
\IEEEauthorblockN{Yifang Zhang}
\IEEEauthorblockA{\textit{Humanoids and Human Centered}\\
\textit{Mechatronics (HHCM) Research Line} \\
\textit{Istituto Italiano di Tecnologia}\\
Genova, Italy \\
yifang.zhang@iit.it}
\and
\IEEEauthorblockN{Nikos G. Tsagarakis}
\IEEEauthorblockA{\textit{Humanoids and Human Centered}\\
\textit{Mechatronics (HHCM) Research Line} \\
\textit{Istituto Italiano di Tecnologia}\\
Genova, Italy \\
nikos.tsagarakis@iit.it}
}

\maketitle

\begin{abstract}
To improve torque capacity and energy efficiency of humanoid ankles, this paper proposes a 2-DoF parallel elastic actuator (PEA). The main novelty of the proposed design lies in its dual-cam, single-gas-spring architecture, which enables torque compensation in both pitch and roll using a shared elastic element, thereby improving structural compactness compared with conventional multi-element compensation schemes. By leveraging parallel gas springs and customized cam modules, the proposed architecture provides dual-axis torque assistance tailored to specific task requirements. The second key contribution is the formulation of a coupled 2-DoF mathematical model that explicitly captures the interdependence between the two compensation units through the shared spring. Based on this model, an optimization-based design framework is developed to synthesize customized cam profiles from prescribed torque references, establishing a systematic link from task requirements to hardware realization. The complete lower-leg CAD integration is presented in detail. Static FEA and kinematic simulations confirm the design's feasibility and torque-relief effectiveness. The results highlight the proposed design as a compact, customizable solution for 2-DoF humanoid ankle torque compensation.

\end{abstract}


\section{Introduction}
Humanoid robots have gained significant traction in both industrial and domestic domains in recent years, due to their enhanced versatility in navigating complex, unstructured, human-centric environments and their anthropomorphic design \cite{tsagarakis2017walk, malik2024intelligent}. Enabling robots to perform real-world tasks requires robots to achieve comparable or even superior performance in terms of load capacity, dynamic capability, and energy efficiency in order to effectively perform tasks encountered in daily life. This, in turn, raises the demand for high-performance actuators and advanced actuation principles. Consequently, improving actuator performance has become a primary research focus in the robotics community.



To achieve high load capacity, conventional actuator designs commonly employ high reduction ratios to increase output torque \cite{Pratt2004}. However, the transmission losses and additional mass associated with high-ratio gearheads inevitably reduce the overall energy efficiency of the actuator, thereby limiting the robot’s operational endurance. Furthermore, the increased reflected inertia introduced by high gear ratios compromises the robot’s dynamic performance and impact resilience.

Driven by advances in motor technology, researchers are increasingly integrating high–torque-density motors with lower-ratio gearboxes, particularly planetary reducers \cite{Xing101007, Seok6880316, Wensing7827048}. This approach enables actuators to maintain comparable load capacity while significantly improving dynamic performance and impact resilience. However, in such configurations, the drive current required to generate the desired torque is typically much higher than in systems with higher reduction ratios. As a result, although the lower reduction ratio mitigates mechanically induced transmission losses, it can significantly increase electrical energy losses which scale with the square of the current. High current during non-dynamic tasks—such as static load holding and manipulation—the system poses challenges for actuator thermal management and reduces the operational endurance of robots.


To balance the dynamic performance and energy efficiency of actuation systems, elastic elements have attracted significant research interest for enhancing actuator performance, owing to their energy storage capability and inherent simplicity as passive components. Depending on their integration topology, such systems are commonly categorized as Series Elastic Actuators (SEAs) and Parallel Elastic Actuators (PEAs). Unlike SEAs, which can exploit stored elastic energy to enhance peak velocity and power output, PEAs are typically employed for torque compensation and for improving the overall energy efficiency of the system.

Various types of elastic elements are utilized in these PEAs, with mechanical tension springs being among the most common. Collins et al. integrated a passive tension spring parallel to the lower leg in a lower-limb exoskeleton to reduce the metabolic cost of human walking \cite{collins2015reducing}. Nan et al. implemented a switchable tension spring mechanism on a single-legged robot, reducing actuator power consumption and enhancing system efficiency \cite{Nan2021}. Meanwhile, Mathews et al. employed a spiral spring with motor-driven stiffness adjustment, allowing the system to store energy during cyclic motions and augment swing amplitude for torque assistance \cite{mathews2022design}.

Compared to conventional metal springs, rubber elastic elements have been increasingly adopted due to their flexibility, lower weight, and high elongation capability. In \cite{Roozing2019}, elastic bands are utilized to store potential energy in order to improve the overall energy efficiency of the system. Similarly, an elastic band is employed to store energy, thereby improving the energy efficiency of a lower-limb assistive device while also providing load compensation for the human joint \cite{Zhang2021Exo-Muscle:Knee}. 

However, incorporating additional elastic elements and their associated integration mechanisms introduces extra weight, which may result in a bulky and heavy system. Therefore, elastic components with high power density and ease of integration are highly desirable for such applications. Gas springs, which often offer higher energy density, compact size, and greater ease of installation, have thus emerged as a promising option. For example, Wang et al. integrated a gas spring in parallel with an electric cylinder in a lower-limb exoskeleton for regulation of the output force \cite{wang2022design}. Similarly, Zhang et al. employed a PEA structure consisting of an eccentric cam and a miniature gas spring in a single-legged robot to achieve torque compensation and power reduction \cite{zhang2024novel}. Furthermore, emerging PEA architectures utilizing magnetic springs have proven effective in boosting efficiency \cite{fu2024energy}.A key advantage of this approach is that it obviates concerns regarding mechanical fatigue and catastrophic failure, which are common limitations of conventional elastic elements.


Nevertheless, most of these solutions that provide load compensation and energy storage require a dedicated standalone elastic element for each actuator. Extending this approach to multi-DoF systems necessitates multiple integrated elastic components, thereby increasing the overall weight and volume of the system. This becomes particularly challenging for joints with multiple degrees of freedom and limited available design space, such as the ankle joint. Previous studies have explored solutions capable of providing bi-articular load compensation using a single elastic element \cite{Eleg}. However, these designs are typically intended for joints arranged in parallel and often require excessive volume to accommodate the elastic element.

To address the challenges of the ankle—one of the most demanding joints in humanoid robots due to its spherical motion, constrained design space, and high torque requirements, this paper introduces the following novel contributions.

\begin{itemize}
    \item  A 2-DoF PEA for ankle joints, employing a dual-cam architecture and a shared gas spring to provide torque compensation for both pitch and roll motions within a constrained installation space.
     \item A comprehensive mathematical framework for the proposed mechanism, extended from a single torque compensation unit to a coupled 2-DoF configuration in which two axes are interconnected through the shared spring force.
     \item An optimization-based design methodology to synthesize customizable cam profiles from specific reference torques, enabling customized compensation behavior and a systematic transition from functional requirements to hardware implementation.
     \item The proposed concept is validated through full lower-leg CAD integration, simulation studies under multiple torque references, and static finite element analysis (FEA), demonstrating the feasibility of the mechanism and the effectiveness of the compensation strategy.
\end{itemize}

This paper is organized as follows: Section II outlines the design principles and provides a detailed mathematical analysis. Section III details the specific mechanical structure of the proposed PEA. Simulation results are presented in Section IV to validate the effectiveness of the torque compensation and the feasibility of specific customization. Finally, Section V concludes the key innovations and contributions.

\section{Methodology}\label{2}
In this section, we begin with a single Torque Compensation Unit (TCU) to introduce its structure, modeling, and mathematical analysis. Based on the selection of moving components and gas spring deployment, this basic unit is extended into three additional structural configurations. Finally, the proposed 2-DoF torque compensation mechanism is presented, which consists of two TCUs sharing the same gas spring.

\begin{figure}[htbp]
\centering{\includegraphics[width=0.7\linewidth]{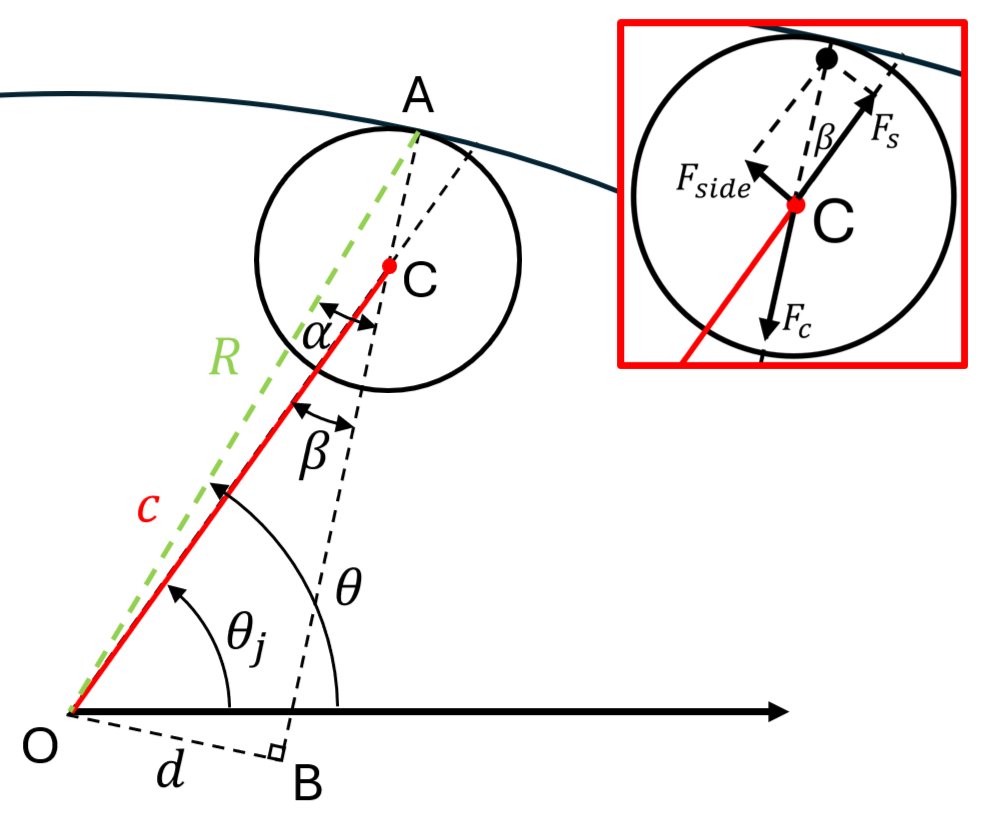}}
\caption{Simplified model of torque compensation unit.}
\label{fig:geo}
\end{figure}

The modeling and analysis of a TCU can be approached starting from 2D geometry. As shown in the polar coordinate in Fig.~\ref{fig:geo}, a roller maintains tangent contact with a non-circular polar curve (the profile of an eccentric cam), where the contact force $F_c$ acts perpendicular to the tangent point A. This force $F_c$ can be decomposed into two components: one acting along the line connecting the pole and the roller center (OC), and the other perpendicular to OC. The component along OC is collinear with the spring force, forming a pair of balanced forces, while the perpendicular component is counterbalanced by forces from other mechanical structures. The line of action of the contact force $F_c(\theta)$, which coincides with the normal AB, does not pass through the pole, but is offset by a distance $d(\theta)$. Consequently, $F_c$ generates a torque $\tau(\theta)$ at the pole with a moment arm $d(\theta)$, enabling either the roller or the cam to rotate. This resultant torque is the compensation torque we intend to achieve and utilize. 

\subsection{Torque Compensation Unit}\label{2A}

Based on this fundamental geometric model, a TCU is realized by aligning the pole with the rotation center of a robotic joint. The cam and roller are respectively mounted on the two links pivoted by this joint. By defining $\theta = 0$ as the zero position, the angle of the line connecting the pole and the roller center (OC) corresponds to the joint rotation angle $\theta_j$. Notably, since the eccentric cam profile is a non-circular curve with a varying radius, the radial variation can translated into the compression stroke of the gas spring through mechanical design, thereby generating the required spring force.

In this parametrically coupled geometric system, given the curve equation $R(\theta)$ and the roller radius $r$, the torque $\tau(\theta)$ at a specific angle $\theta$ can be derived by determining the contact force $F_c(\theta)$ and the moment arm $d(\theta)$ respectively. In a polar coordinate system, the angle $\alpha(\theta)$ between the polar radius OA and the normal line AB at any point on a smooth curve is governed by the following relationship.
\begin{equation}\label{alpha}
    \alpha(\theta) =|\arctan\frac{R'(\theta)}{R(\theta)}|
\end{equation}
where $R'(\theta)$ represents the first derivative of $R(\theta)$.

Consequently, the moment arm $d(\theta)$ can be obtained through the trigonometric relationship.
\begin{equation}\label{d}
    d(\theta) = R(\theta)\sin(\alpha(\theta))
\end{equation}

Regarding the contact force $F_c(\theta)$, since its component is in equilibrium with the spring force $F_s(\theta)$, the following condition must be satisfied.
\begin{equation}\label{Fc}
    F_c(\theta) = \frac{F_s(\theta)}{\cos \beta(\theta)}
\end{equation}
where $\beta(\theta)$ denotes the angle between the segment OC and the normal line, which can be derived from the following trigonometric relationship.
\begin{equation}\label{beta}
    \cos\!\bigl(\beta(\theta)\bigr) = \frac{\sqrt{\,c(\theta)^{2} - R(\theta)^{2}\,\sin^{2}\!\alpha(\theta)\,}}{c(\theta)}
\end{equation}
Here, $c(\theta)$ represents the length of the line segment OC at angle $\theta$, and is obtained by applying the Law of Cosines to triangle OBC.
\begin{equation}\label{c}
    c(\theta) = \sqrt{R(\theta)^2 + r^2 - 2R(\theta)r\cos\alpha(\theta)}
\end{equation}

The spring force of the gas spring originates from its compression stroke, which is determined by the variation in the distance between the pole and the roller center, denoted as $\Delta c(\theta)$.

From this, $\Delta c(\theta)$ representing the spring compression as the joint rotates from the zero position to $\theta_j$ is given by
\begin{equation}\label{deltac}
    \Delta c(\theta) = c(0) - c(\theta) = R(0) - r - c(\theta)
\end{equation}

Taking into account the initial spring force $F_{si}$ and the pre-load displacement $c_p$, the instantaneous spring force $F_s(\theta)$ can be formulated as
\begin{equation}\label{Fs}
    F_s(\theta) = F_{si} + k(\Delta c(\theta)+c_p)
\end{equation}
where $k$ denotes the stiffness of the spring.

By substituting the expression for $F_s(\theta)$ into \eqref{Fc}, the contact force $F_c(\theta)$ is identified. Combined with the moment arm $d(\theta)$ obtained from \eqref{d}, the resulting compensation torque $\tau(\theta)$ at angle $\theta$ can be solved as
\begin{equation}\label{tau}
    \tau(\theta) = F_c(\theta) \cdot d(\theta)
\end{equation}

However, in practical applications, the parameter $\theta$ is not directly measurable. Instead, the robotic joint position $\theta_j$ is readily obtained via sensors and is essential for position control. Consequently, our concern is the relationship between the torque and the joint angle $\theta_j$. From the geometric configuration, the following angular relationship can be derived.
\begin{equation}\label{thetaj}
    \theta = \theta_j + \beta - \alpha
\end{equation}

By substituting this into the expression for $\tau(\theta)$, the torque as a function of the joint angle, $\tau(\theta_j)$, can be determined.

\begin{figure}[htbp]
\centering{\includegraphics[width=0.8\linewidth]{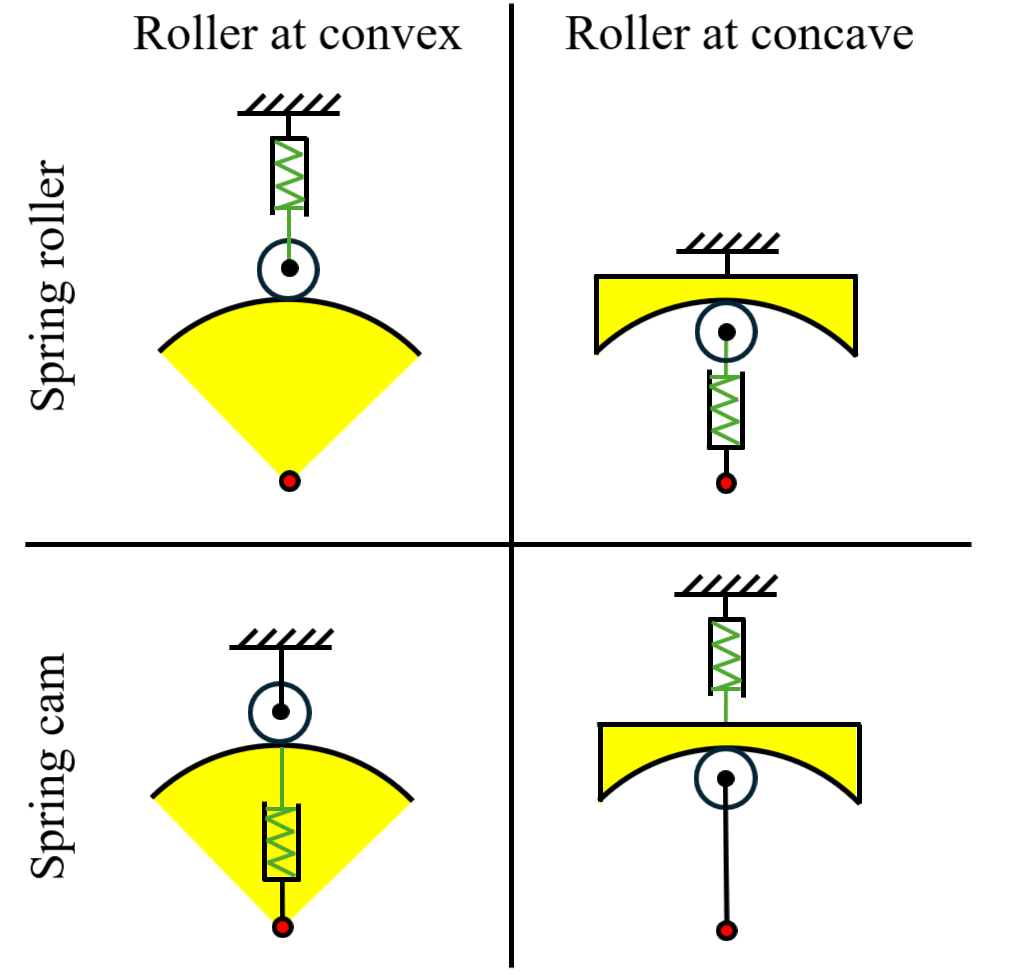}}
\caption{Four configurations of torque compensation unit.}
\label{fig:config}
\end{figure}

Furthermore, the TCU can be realized in four different forms in Fig.~\ref{fig:config}, depending on the roller's placement relative to the curve's concavity (convex vs. concave) and the allocation of the spring force to either the cam or the roller assembly. In the figure, the roller is shown in white, the cam in yellow, and the gas spring in green. The center of rotation is indicated in red and remains positioned on the concave side of the cam profile. The simplified geometric model shown in Fig.~\ref{fig:geo} corresponds to the configuration where the roller is on the concave side and the gas spring acts on the roller, as illustrated in the top-right subplot of Fig.~\ref{fig:config}. Following a mathematical analysis analogous to the TCU described above, the torque expressions for the other three configurations can be similarly derived.

\subsection{2-DoF Torque Compensation Mechanism}\label{2B}
As illustrated by the four configurations in Fig.~\ref{fig:config}, the TCU with the roller positioned on the concave side exhibits a more compact structure given the same cam profile radius. Furthermore, based on the surface area of the cam (indicated by the yellow regions), it is evident that—assuming a homogeneous solid cam—placing the roller on the concave side significantly reduces the volume of material compared to the convex configuration, resulting in a lighter overall assembly.

Specifically, for the humanoid ankle joint targeted in this study, the compact concave-side roller configuration is better suited to the highly constrained spatial environment. Furthermore, to facilitate customizable compensation tailored to specific torque requirements, cams must be easily detachable and replaceable. This requires placing the cams away from the structurally complex 2-DoF rotational axes of the ankle. To meet this requirement, we adopted a layout with two cams positioned at the distal ends and two rollers on the proximal side. To further enhance compactness, a single gas spring is shared between the two TCUs, exerting force on both rollers. Consequently, we selected the configuration where the rollers are on the concave side with the spring acting between them. This is implemented by connecting two such TCUs in a back-to-back series, forming the 2-DoF torque compensation mechanism illustrated in Fig. 3.

\begin{figure}[htbp]
\centering{\includegraphics[width=0.8\linewidth]{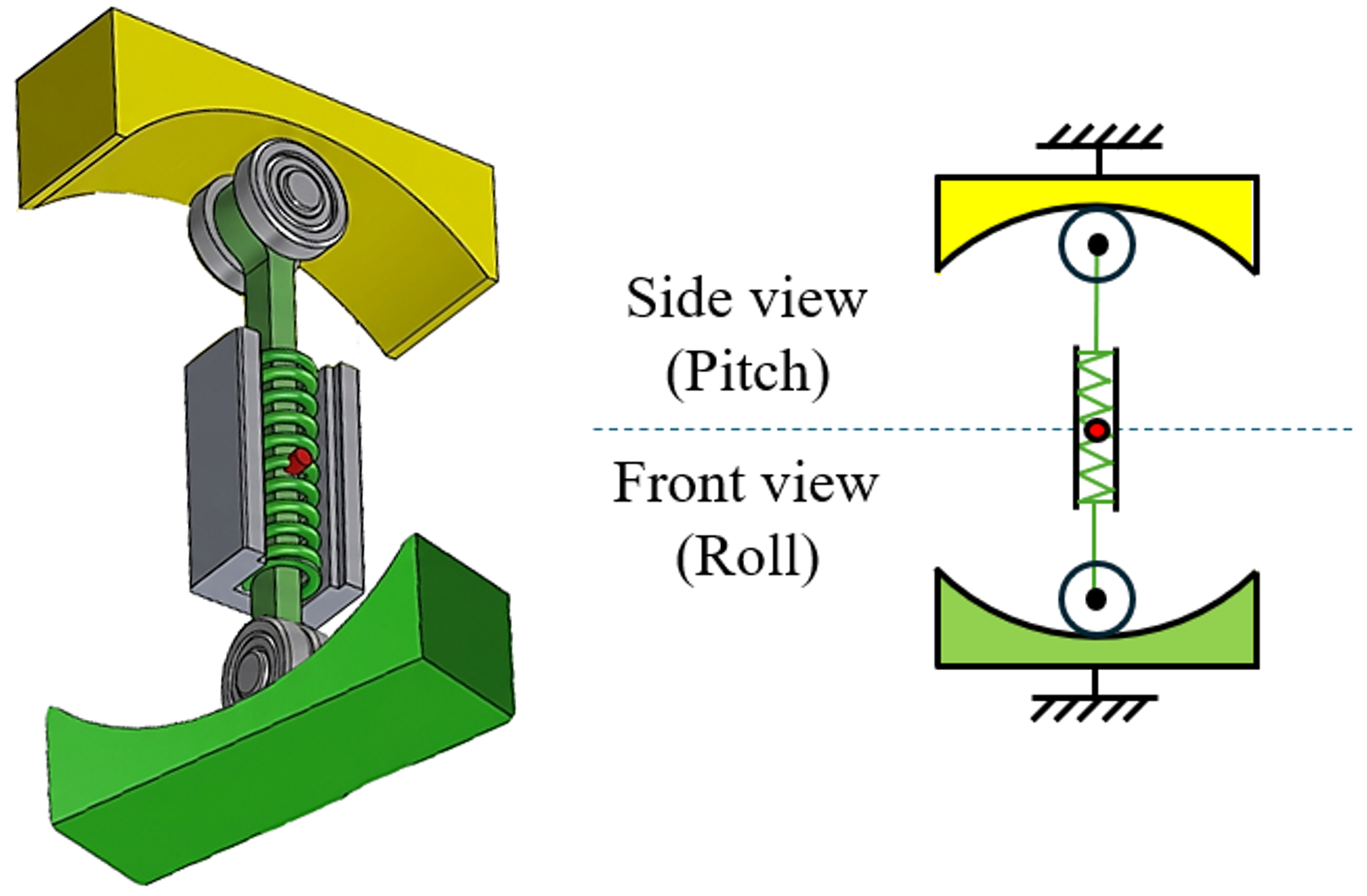}}
\caption{2-DoF torque compensation mechanism of two orthographic views combined.}
\label{fig:tcm}
\end{figure}

It should be noted that two DoFs of the ankle are orthogonal. For the sake of clear representation in a two-dimensional schematic, Fig.~\ref{fig:tcm} is constructed by combining the upper and lower TCUs from two orthographic views rotated $90^\circ$ relative to each other. Within the ankle joint, one TCU provides the compensation torque for the pitch motion, while the other handles the roll motion. The rotational axes of these two DoFs can be either coplanar or configured with offset as skew lines.

The mathematical model of the 2-DoF torque compensation mechanism is essentially a combination of two individual TCU models. The key distinction lies in the shared gas spring: the total spring compression is the sum of the displacements from both TCUs. Consequently, the resulting spring force—governed by this aggregate compression—acts simultaneously on both rollers, thereby creating a mathematical coupling between the two TCU models.
\begin{equation}\label{fs2}
    F_s(\theta_1, \theta_2) = F_{si} + k(\Delta c_1(\theta_1) + \Delta c_2(\theta_2) + c_p)
\end{equation}
Where $\theta_1$ and $\theta_2$ denote the angles of the contact points on the cam profiles, respectively (as distinguished from the joint angles). The terms $\Delta c_1(\theta_1)$ and $\Delta c_2(\theta_2)$ represent the individual contributions of each TCU to the spring compression; consequently, the gas spring force is interdependent on both $\theta_1$ and $\theta_2$. Consequently, the output torques of two TCYs are interdependently governed by both $\theta_1$ and $\theta_2$.
\begin{equation}\label{tau2}
\begin{aligned}
    \tau_1(\theta_1, \theta_2) = F_s(\theta_1, \theta_2) \cdot d_1(\theta_1) \\
    \tau_2(\theta_1, \theta_2) = F_s(\theta_1, \theta_2) \cdot d_2(\theta_2)
\end{aligned}
\end{equation}

\subsection{Optimization-based Torque Compensation}\label{2C}
In Section~\ref{2A}, we demonstrate the fundamental principles of torque generation in the TCU and provide a detailed mathematical analysis. Furthermore, we aim for the torque generated by the TCU to match or closely approximate a pre-defined reference torque to meet specific task requirements. Since our torque derivation originates from an unknown polar curve $R(\theta)$, it is necessary to first define its functional form. To ensure mathematical simplicity and sufficient approximation power while maintaining computational tractability for subsequent optimization, we represent $R(\theta)$ using an $m$-th order polynomial with a set of unknown parameters $u_i$.
\begin{equation}\label{rtheta}
    R(\theta) = R_0 + \sum_{i=1}^{m} u_i \theta^i
\end{equation}
where $R_0$ is a constant, indicating the initial radius at the zero position.

We formulate an optimization problem to solve for the unknown parameters of $R(\theta)$, thereby determining its explicit expression and minimizing the discrepancy between the reference torque and the compensation torque. Given a prescribed reference torque function $\tau_{ref}(\theta_j)$ and an $n$-dimensional position vector $\mathbf{\Theta}$ uniformly sampled within the range $(\theta_{min}, \theta_{max})$, these are substituted into the TCU model alongside $R(\theta)$. This yields a corresponding set of $n$-dimensional joint position vector $\mathbf{\Theta_j}$ and compensation torque vector $\mathbf{T}$. By calculating the reference torque equation $\tau_{ref}(\theta_j)$ at the computed $\mathbf{\Theta_j}$, we obtain the reference torque vector $\mathbf{T}_{ref}$. The residual matrix $\mathbf{E}$, representing the difference between $\mathbf{T}_{ref}$ and $\mathbf{T}$, characterizes the approximation error. The sum of squared errors (SSE) is defined as the objective function. The optimization is subject to the constraint that the spring compression must not exceed its maximum stroke $s$. Furthermore, to prevent geometric undercut and ensure that the roller can perfectly follow the cam profile, a curvature constraint must be satisfied. Specifically, the minimum radius of curvature of the cam profile, $\rho(\theta)$, must be strictly greater than the roller radius $r$ at all contact points. The structure of this optimization problem is summarized as follows.

\begin{equation}
\begin{aligned}
    \min_{\mathbf{u}} \quad & J = \mathbf{E}^T \cdot \mathbf{E} \\
    \text{s.t.} \quad & 0 \le \Delta c + c_p \le s \\
    & \rho(\theta) > r, \quad \forall \theta \in (\theta_{min}, \theta_{max})
\end{aligned}
\end{equation}
with\
\begin{equation}\label{errorvec}
    \mathbf{E} = |\mathbf{T}_{ref} - \mathbf{T}|
\end{equation}
where the i-th elements of the torque vectors and position vectors satisfy
\begin{equation}\label{tauvec}
    \tau_{ref,i} = \tau_{ref}(\theta_{j,i}),\quad
    \tau_{i} = \tau(\theta_i),\quad
    \forall i \in \{1, \dots, n\}
\end{equation}

Substituting the optimized parameters $\mathbf{u}$ into the expression for $R(\theta)$ yields the optimal profile for torque compensation, contributing to the subsequent mechanical design of the cam.

This study extends the torque compensation approach presented in \cite{zhang2024novel} from a single DoF to a 2-DoF system using a single gas spring unit. Accordingly, the optimization problem is reformulated. Despite this extension, the overall structure of the optimization formulation for the TCU remains analogous. However, due to the mathematical coupling between the two TCUs, both the prescribed reference torques $\tau_{r1}(\theta_p, \theta_r), \tau_{r2}(\theta_p, \theta_r)$ and the output compensation torques $\tau_1(\theta_1, \theta_2), \tau_2(\theta_1, \theta_2)$ are bivariate functions of both the pitch and roll degrees of freedom. The constraints regarding the spring stroke and the radius of curvature are similarly extended from the constraints of 1-DoF situation. Therefore, the optimization problem is formulated as follows.
\begin{equation}\label{opt2}
\begin{aligned}
\min_{u_1,u_2} \quad & J = \mathbf{E_1}^T\cdot \mathbf{E_1} +  \mathbf{E_2}^T\cdot \mathbf{E_2}\\
\textrm{s.t.} \quad &  0 \leq \Delta c_1 + \Delta c_2 + c_p \leq s\\
& \rho(\theta_1) > r_1, \quad \forall \theta_1 \in (\theta_{1min}, \theta_{1max})\\
& \rho(\theta_2) > r_2, \quad \forall \theta_2 \in (\theta_{2min}, \theta_{2max})
\end{aligned}
\end{equation}
with
\begin{equation}\label{error2vec}
    \mathbf{E}_1 = |\mathbf{T}_{r1} - \mathbf{T}_1|,\quad \mathbf{E}_2 = |\mathbf{T}_{r2} - \mathbf{T}_2|
\end{equation}
where the elements in the i-th row and k-th column of the corresponding torque matrices satisfy

\begin{equation}
\begin{aligned}
    \tau_{r1,ik} &= \tau_{r1}(\theta_{p,i}, \theta_{r,k}), \quad &\tau_{r2,ik} &= \tau_{r2}(\theta_{p,i}, \theta_{r,k}) \\
    \tau_{1,ik} &= \tau_1(\theta_{1,i}, \theta_{2,k}), \quad &\tau_{2,ik} &= \tau_2(\theta_{1,i}, \theta_{2,k}) \\
    &\forall i \in \{1, \dots, n_1\}, \quad &\forall k &\in \{1, \dots, n_2\}
\end{aligned}
\end{equation}
where $n_1$ and $n_2$ depend on the required discretization density and the angular reachability range. Analogous to \eqref{thetaj} in the model of 1-DoF TCU, the cam profile angles $\theta_1$ and $\theta_2$ in this context are not identical to the joint angles $\theta_p$ and $\theta_r$.

With the parameters $\mathbf{u}_1$ and $\mathbf{u}_2$ identified, the geometric profiles $R_1(\theta_1)$ and $R_2(\theta_2)$ for cams of both degrees of freedom are fully defined, providing the essential data for the following deaign.

\section{System Design}\label{3}
The optimization results from Section~\ref{2C} are translated into the physical design of the torque compensation mechanism, encompassing the cam profile generation and the selection of the commercial parts including gas spring and rollers. The design of the PEA further includes the specification of the motors and transmission systems. Finally, the complete CAD model of the lower leg, integrated with the 2-DoF ankle joint, is designed and assembled in CREO.

\subsection{Cam Design workflow}\label{3A}
\begin{figure}[htbp]
	\centering
	\includegraphics[width=1\linewidth]{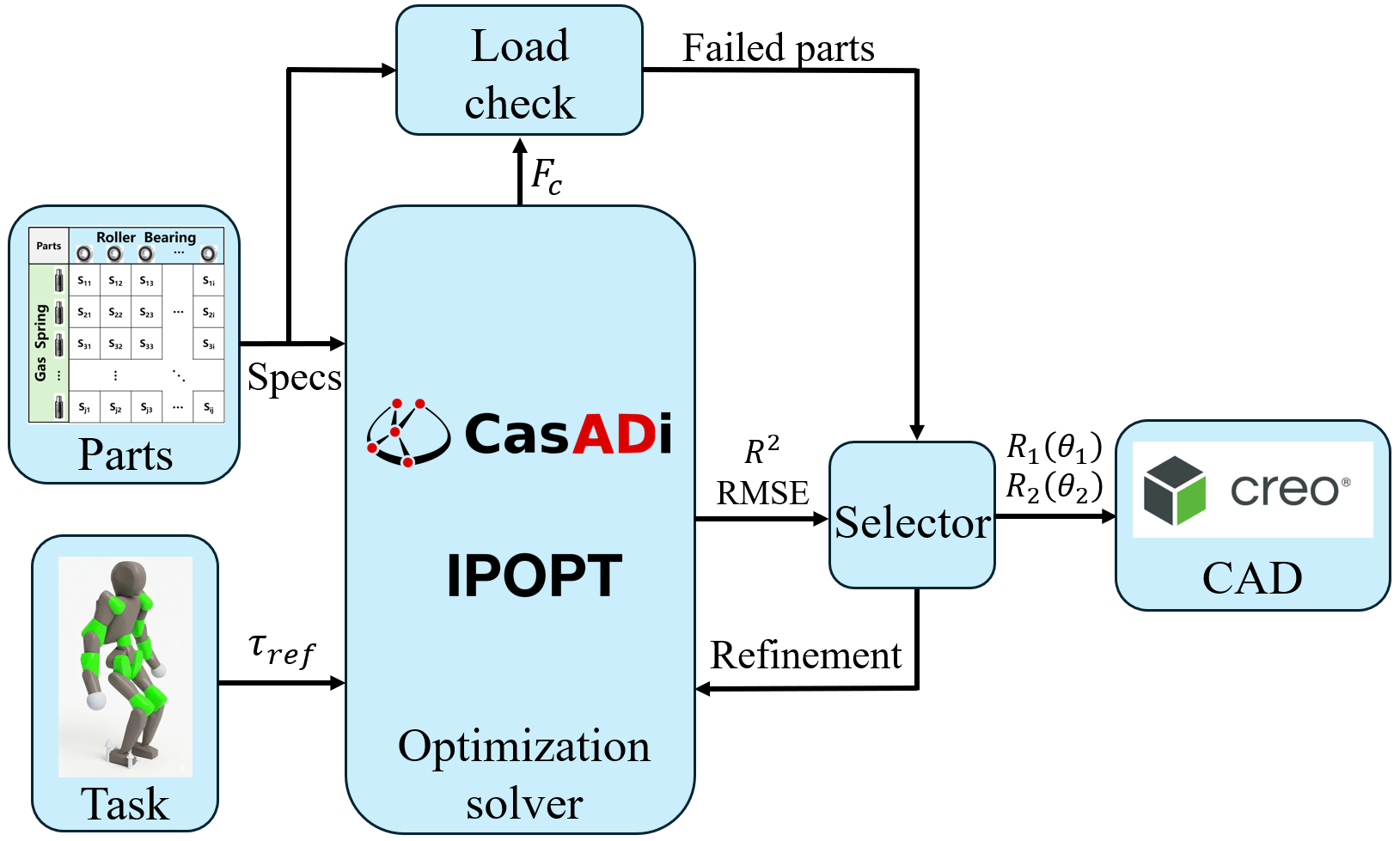}
\caption{Optimization-based workflow for task-oriented synthesis of cam profiles and component selection.}
\label{fig:flow}
\end{figure}

The proposed optimization-based design workflow, as illustrated in Fig.~\ref{fig:flow}, establishes an integration from functional requirements to physical implementation. The process originates from the extraction of reference torque trajectories based on specific robotic ankle task scenarios. These torque reference, along with a structured parts library specifications of the off-the-shelf components, including the force and stroke of various gas springs, as well as the radius of roller bearings of different sizes, constitute the foundational design inputs for the workflow. These design parameters are integrated into the formulated optimization problem, which is subsequently solved using the IPOPT solver within the CasADi framework \cite{CasADi2018}. This core module solves for the optimal parameters of the cam profiles while simultaneously evaluating performance metrics and contact forces ($F_c$). Furthermore, the optimization module transmits the calculated contact forces to a roller bearing verification module for load check. The primary objective of this stage is to evaluate whether the resultant loads exceed the dynamic or static load capacities of the candidate bearings, incorporating a specific overload factor (safety margin) to ensure reliability. Configurations that fail to meet the requirements are flagged and relayed to the subsequent selector module as reference. After filtering out the unqualified candidates, the selector compares the remaining solutions based on their performance metrics to identify the optimal part combination and its corresponding analytical profile expression. However, if the results fail to converge or the performance metrics are insufficient, a feedback loop is triggered from the selector to the solver module for refinement, which involving re-selecting initial parameters or appropriately scaling down the reference torque. Finally, the synthesized design parameters are imported into Creo software, facilitating the precise 3D CAD construction and mechanical integration of the proposed torque compensation mechanism.

\subsection{Actuator and Transmission Design}
The lower-leg design employs a parallel actuation scheme to drive the 2-DoF ankle joint. This configuration reduces the peak torque requirements of each individual motor, thereby enabling the use of lighter and more compact actuators.

In conventional RSS parallel four-bar linkage schemes, actuators typically rely on planetary or harmonic reducers to amplify torque \cite{tsagarakis2017walk}. To further reduce the mass of the lower leg and simplify actuator architecture, we adopt a ball screw transmission, thereby eliminating the need for auxiliary reducers. The intrinsic torque multiplication capability of the ball screw reduces the overall mass of the transmission chain while improving the system’s power density and mechanical efficiency. Furthermore, the use of a ball screw shifts the center of mass toward the proximal region of the humanoid system, which improves dynamic performance by reducing distal inertia and the resulting dynamic loads during locomotion.


In our design, we employ two F50-13 motors from Wuji Ltd, featuring a loaded speed of 6800 rpm and a rated torque of 1.27 Nm (sustained for 10 seconds). The SOMANET Circulo 7 from Synapticon GmbH is utilized as the motor controller, which strikes an optimal balance between high-performance requirements and spatial compactness. 

\begin{figure}[htbp]
	\centering
	\includegraphics[width=0.9\linewidth]{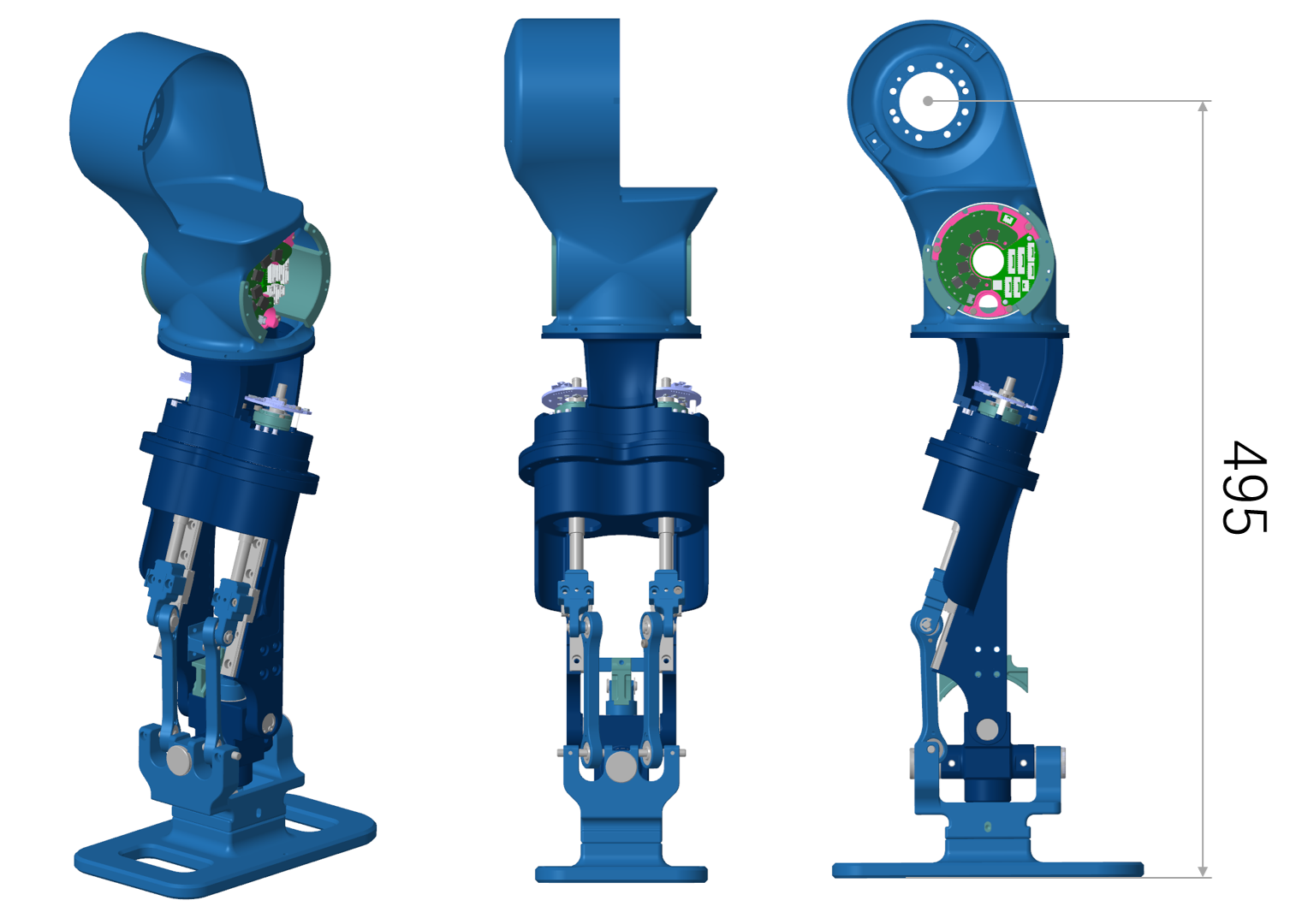}
\caption{CAD of the lower leg.}
\label{fig:cad}
\end{figure}

The full-scale CAD assembly of the lower leg, spanning from the knee joint to the foot, is illustrated in Fig.~\ref{fig:cad}. The screw nut is embedded within the rotor, driving a BSSC-1004 ball screw to execute prismatic motion. This motion actuates the underlying crank-slider mechanism, thereby driving the ankle rotation. The dual parallel ball screw transmissions enable the ankle to perform 2-DoF movements in both pitch and roll. The overall height, comprising the length of the shank and the thickness of the foot, totals 495 mm. The total mass of the lower leg with foot is 3.56 kg. The dimension and mass are consistent with the lower leg proportions of a full-scale 1.7 m, 65 kg humanoid robot. For a robot of this size and weight, based on foot area and safety margin, we set the ankle torque requirements to be 127 Nm for pitch motion and 70 Nm for roll motion. The selection of the main drivetrain, including the motor and ball screw, was based on kinematic modeling calculations to meet the torque requirements. Due to space limitations, detailed explanations are omitted here.

\begin{figure}[htbp]
	\centering
	\includegraphics[width=0.9\linewidth]{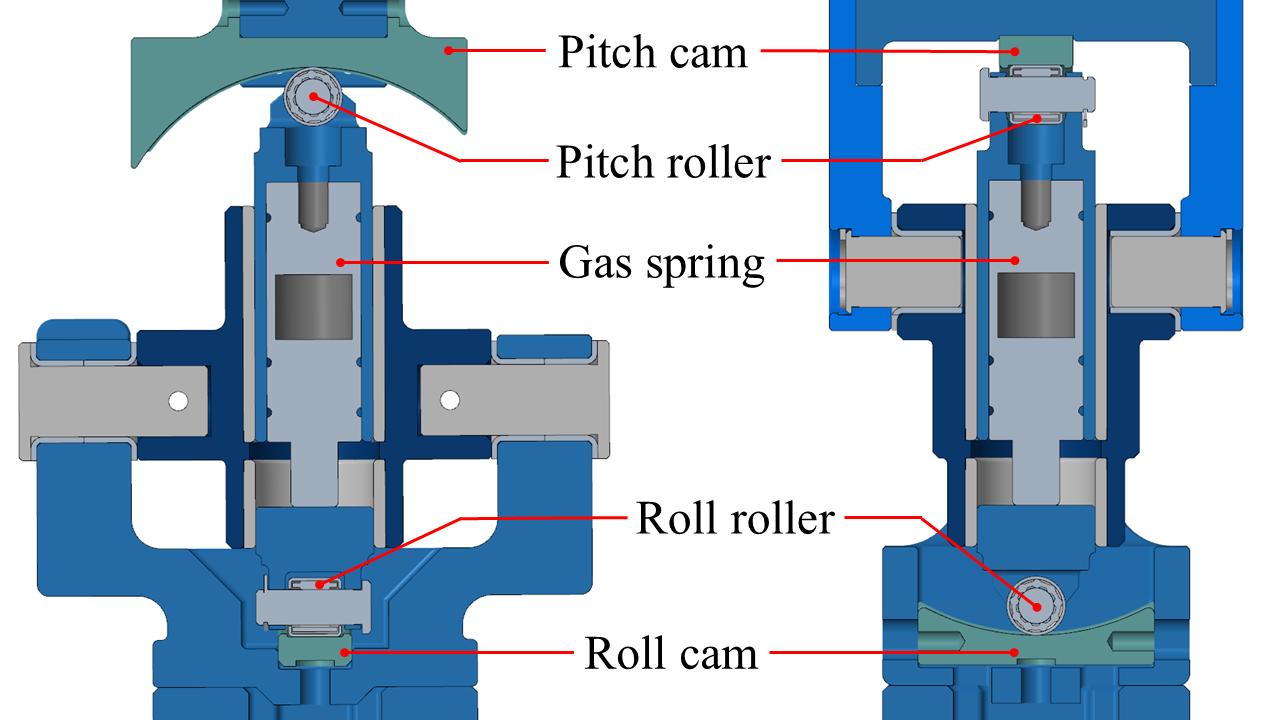}
\caption{Section views of the ankle joint.}
\label{fig:sv}
\end{figure}

The structural details of the torque compensation mechanism are presented in the section views in Fig.~\ref{fig:sv}, which consists of a side view and a front view. Two customizable cams are positioned at the top and bottom, respectively, flanking a pair of rollers that share a common gas spring. HK0609FM needle roller bearings with a 10 mm diameter were selected as rollers. Both their dynamic and static load ratings exceed 2500 N, allowing the mechanism to robustly handle the high-load scenarios typically encountered at the ankle joint. Two gas spring with similar size are selected from MISUMI. The GSU70-7-YW provides a force ranging from 700 N to 868 N with a 7 mm stroke, whereas the MGSN16-10 delivers 1000 N to 1405 N over a 10 mm stroke. To avoid the significant volume and mass of conventional linear bearings, the spring is housed within a self-lubricating graphite bushing, which provides lateral support while reducing frictional losses. As the ankle performs pitch or roll movements, the corresponding rollers travel along their respective cam profiles, compressing the spring. Simultaneously, the spring reciprocates vertically within the bushing under the compression from both ends.

\section{Simulation Validation}\label{4}
In this section, we implement the cam design for several prescribed torque references, leveraging the mathematical model from Section~\ref{2} and the optimization-based design workflow established in Section~\ref{3}. Furthermore, we evaluate and analyze the compensation performance of the resulting dual-output torques generated by the mechanism. The designed cams originated from one set of the prescribed torque references are analyzed with static FEA to validate the strength.

\subsection{Torque Compensation Validation}\label{4A}
In typical standing tasks, the ankle joint torque tends to increase as the joint deviates from its neutral position. Therefore, three sets of sinusoidal functions are initially selected as torque references to perform a preliminary validation of the proposed method's effectiveness. Given the robot's nominal mass of approximately 65 kg and the foot dimensions (200 $\times$ 110 mm), the peak torque for a single foot—assuming an overload factor of 2—is calculated as 127 Nm for the pitch motion and 70 Nm for the roll motion. Assuming a peak output of 10$\%$ for the compensation ratio, which defines the maximum magnitude of the sinusoidal references, the following two sets of expressions are selected as the torque references for our validation.
\begin{equation}\label{tauref1}
    \tau_{r1} = 12.7 \sin(\theta_p) ,\quad
    \tau_{r2} = 7 \sin(\theta_r)
\end{equation}

\begin{equation}\label{tauref2}
\begin{aligned}
    \tau_{r1} = 10 \sin(\theta_p) + 2.7 \cos(\theta_r) \\
    \tau_{r2} = 1.5 \cos(\theta_p) + 5.5 \sin(\theta_r)
\end{aligned}
\end{equation}

These two sets of references represent the decoupled and coupled scenarios for the target pitch and roll torques, respectively. Furthermore, for a single-leg load test of the lower leg, assuming a 4 kg payload is mounted at the top, the ankle torque at various postures on flat ground—given a shank length of 400 mm and  gravitational constant of $10$ N/kg — satisfy the following equilibrium conditions:
\begin{equation}\label{tauref13}
    \tau_{r1} = 16 \sin(\theta_p) ,\quad
    \tau_{r2} = 16 \cos(\theta_p) \sin(\theta_r)
\end{equation}

\begin{figure}[htbp]
    \centering
    \includegraphics[width=0.9\columnwidth]{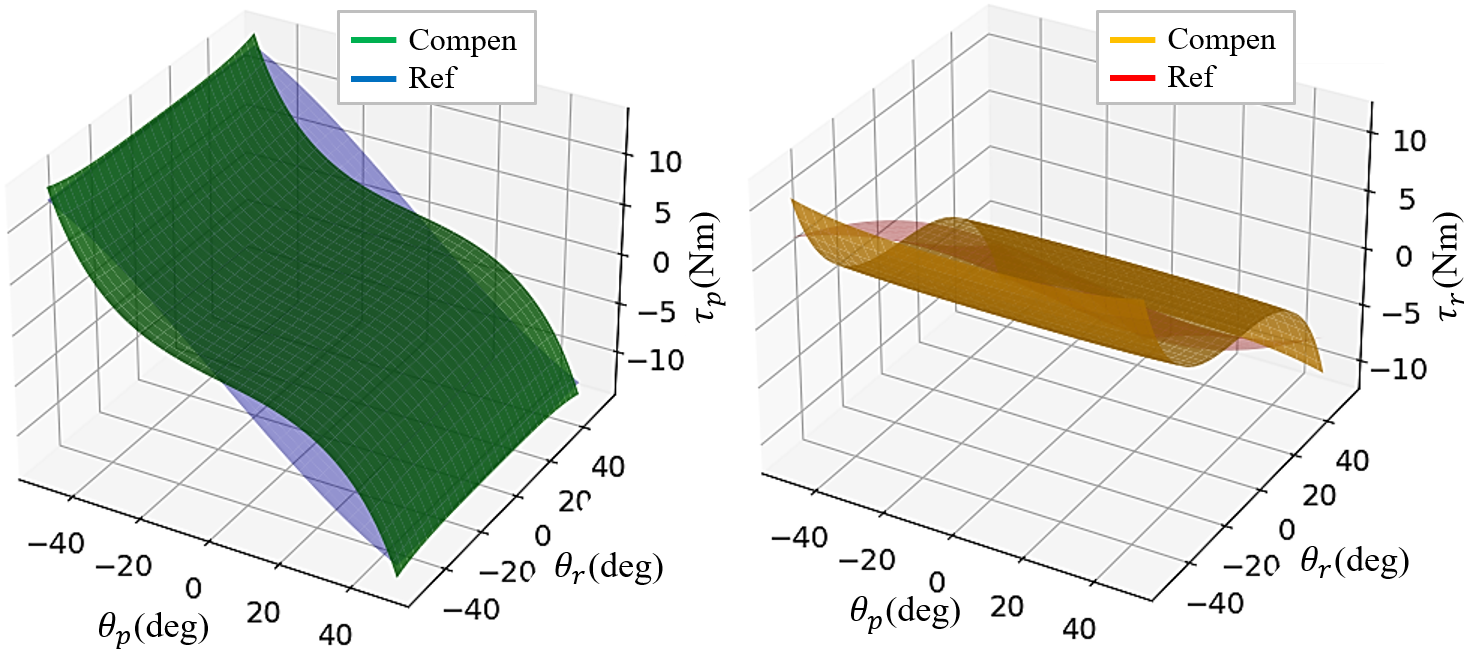}
    \caption{Compensated and reference torque.}
    \label{fig:sm1}
\end{figure}

\begin{figure}[htbp]
    \centering
    \includegraphics[width=0.9\columnwidth]{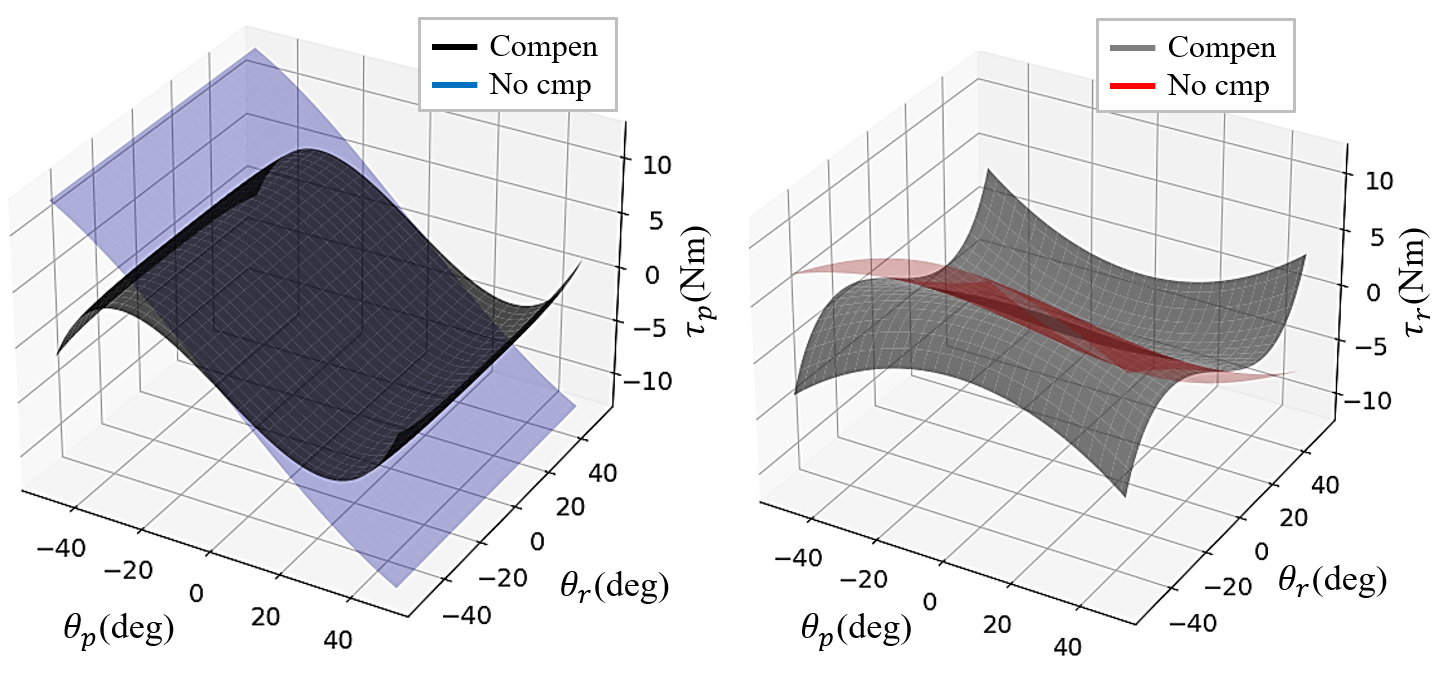}
    \caption{Ankle torque provided by motors.}
    \label{fig:sm2}
\end{figure}

Considering both computational efficiency and optimization accuracy, a fourth-order polynomial provides sufficient flexibility to approximate the sinusoidal references used in this study, where $m$ is taken as 4 in \eqref{rtheta}. Following the design workflow described in Section~\ref{3A}, the optimization was performed for three reference sets with two springs. As presented in Table~\ref{tab1}, torque fitting is almost perfectly achieved under decoupled reference, whereas the performance is relatively constrained under fully coupled scenarios. This discrepancy is attributed to the mathematical coupling between the two TCUs, where an inherent interdependence exists between the pitch and roll compensation torques. A comparison between the two springs reveals that the MGSN spring, which provides a higher elastic force, yields superior fitting results and a more significant reduction in torque. Although the GUS spring exhibits slightly lower compensation performance, its reduced spring force mitigates the structural stress on components. Furthermore, the higher magnitude of the pitch reference generates a dominant gradient contribution within the objective function, naturally prioritizing its convergence. Consequently, under identical conditions, the fitting accuracy and compensation performance for pitch surpass those of roll. The torque plots of 4 kg load reference are shown in Fig.~\ref{fig:sm1}, showing the fitting performance and the reduction in ankle torque provided by motor. The Fig.~\ref{fig:sm2} corresponds to the average torque drop and peak torque drop in Table~\ref{tab1}, relaxing the requirements for motor selection.
\vspace{3mm}

\begin{table}[htbp]
    \centering

    \caption{Performance with different springs and reference.}
    \vspace{2mm}

    \resizebox{0.5\textwidth}{!}{
        \begin{tabular}{ll cccccc}
            \toprule
            \textbf{Spring} & \textbf{Reference} & \multicolumn{2}{c}{$R^2$} & \multicolumn{2}{c}{\textbf{Avg Drop}} & \multicolumn{2}{c}{\textbf{Peak Drop}} \\
            \cmidrule(lr){3-4} \cmidrule(lr){5-6} \cmidrule(lr){7-8}
            & & $\tau_p$ & $\tau_r$ & $\tau_p$ & $\tau_r$ & $\tau_p$ & $\tau_r$ \\
            \midrule
            
            \multirow{3}{*}{\shortstack[l]{GSU70-\\7-YW}} 
                            & Decoupled & 0.99 & 0.99 & 79.0\% & 58.7\% & 81.0\% & 64.6\% \\
                            & Coupled   & 0.87 & 0.67 & 68.4\% & 52.1\% & 69.4\% & 51.6\% \\
                            & Payload & 0.81 & 0.71 & 56.0\% & 45.5\% & 60.3\% & 50.5\% \\
            \midrule
            
            \multirow{3}{*}{\shortstack[l]{MGSN\\16-10}}  
                            & Decoupled & 0.99 & 0.99 & 97.7\% & 95.3\% & 95.2\% & 90.3\% \\
                            & Coupled   & 0.94 & 0.92 & 78.6\% & 76.0\% & 76.1\% & 73.8\% \\
                            & Payload & 0.99 & 0.96 & 96.0\% & 81.9\% & 91.9\% & 65.6\% \\
            \bottomrule
        \end{tabular}
        \label{tab1}
    } 
\end{table}

\begin{figure}[htbp]
	\centering
	\includegraphics[width=0.65\linewidth]{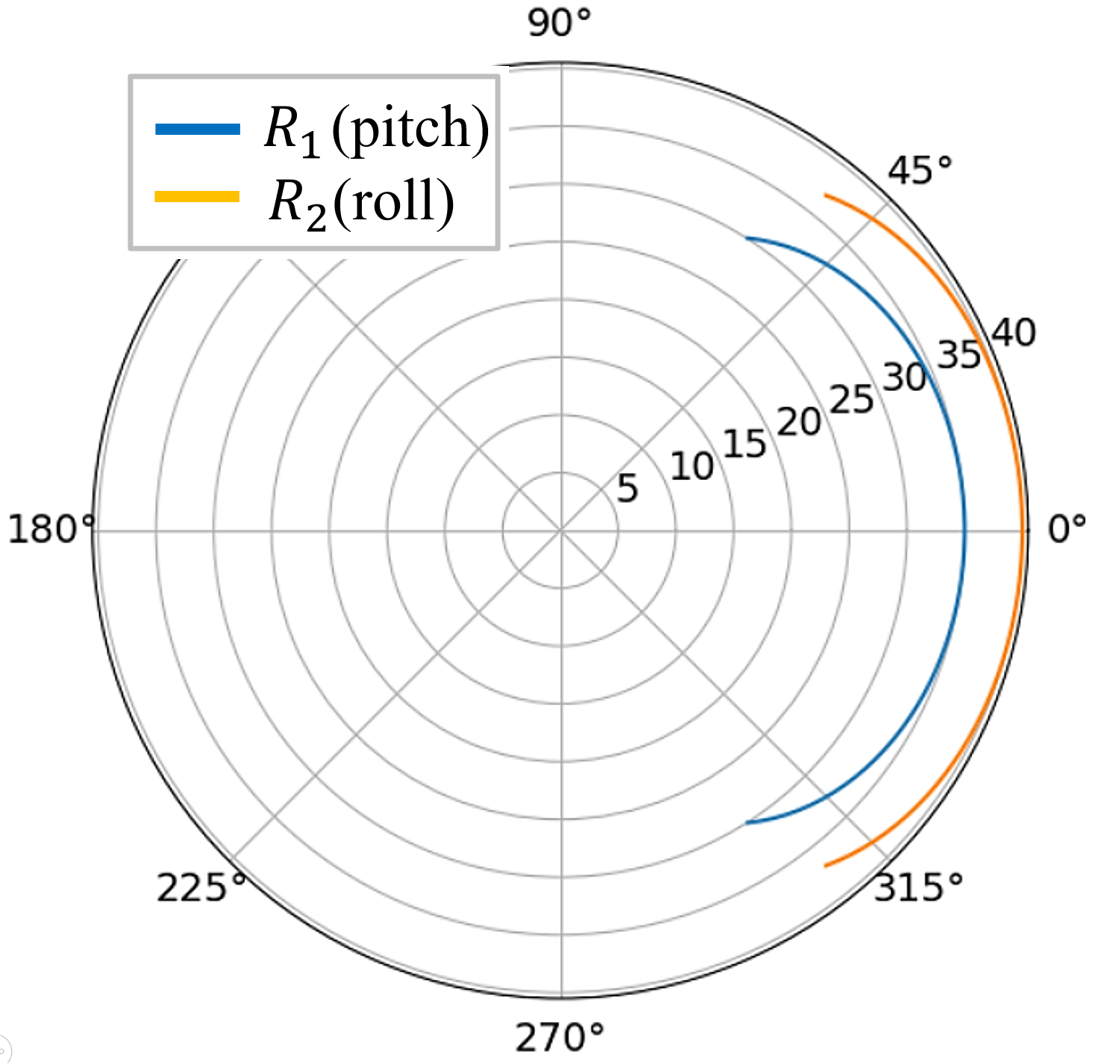}
\caption{Optimized cam profiles of pitch and roll axes derived from 4 kg payload reference with spring GSU70-7-YW.}
\label{fig:r}
\end{figure}

Fig.~\ref{fig:r} illustrates the optimal profiles derived with the 4 kg payload reference. The corresponding mathematical expressions are formulated as follows.
\begin{equation}\label{r1r2}
\begin{aligned}
    R_1(\theta_1) = {} & -2.6774 \theta_1^4 - 2.2700e^{-16}\theta_1^3 \\
                       & - 2.2639\theta_1^2 + 5.8127e^{-15}\theta_1 + 35 \\[1ex]
    R_2(\theta_2) = {} & -1.7344e^{-2}\theta_2^4 - 7.0267e^{-16}\theta_2^3 \\
                       & - 5.4511\theta_2^2 + 1.1315e^{-15}\theta_2 + 40
\end{aligned}
\end{equation}

\subsection{Finite Element Analysis and Validation}\label{4B}
\begin{figure}[htbp]
    \centering

    \subfloat[Pitch cam]{%
        \includegraphics[width=0.8\columnwidth]{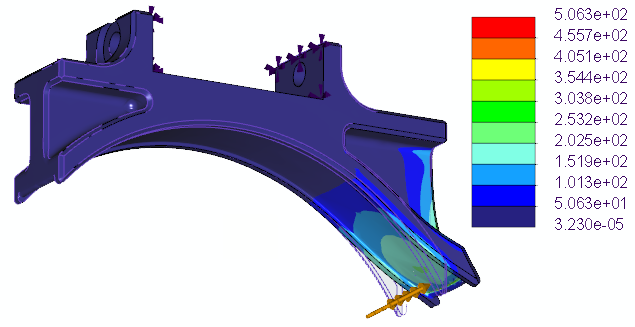}
        \label{fig:fea1}
    }
    \hfill
    \subfloat[Roll cam]{%
        \includegraphics[width=0.8\columnwidth]{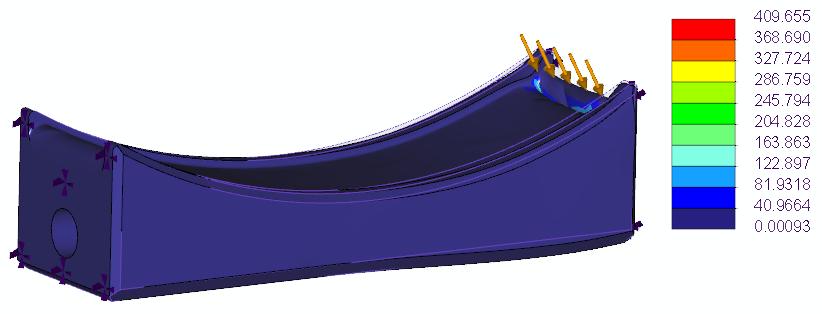}
        \label{fig:fea2}
    }

    \caption{Static FEA of the pitch and roll cam modules under peak loading conditions.}
    \label{fig:fea}
\end{figure}

Based on the optimal payload profiles derived from \eqref{r1r2} with spring GSU70-7-YW, the customizable cam modules for both pitch and roll axes were designed. A static FEA was conducted within the Creo simulation environment to evaluate the two cam modules of 7075-T6 aluminum alloy under peak loading conditions with the 4 kg load reference, as illustrated in Fig.~\ref{fig:fea}. Although finite element analysis (FEA) indicates that the maximum subsurface von Mises stress (approximately 500 MPa) approaches the tensile yield strength of the 7075-T6 aluminum alloy, the single-occurrence static loading condition in this application effectively eliminates the risk of contact fatigue failure induced by alternating loads. Furthermore, based on Hertzian contact theory, this peak stress is highly localized within a microscopic subsurface region; under the strong constraint of the surrounding large-scale elastic matrix (a state of triaxial compression), it remains far below the true critical contact pressure required to induce macroscopic plastic indentation on the surface. Consequently, this highly localized stress concentration will not lead to macroscopic yielding or structural failure of the material, keeping it well within the acceptable safety margins for engineering design.




\section{Conclusions and Future Work}
To enhance the torque capacity and energy efficiency of humanoid robotic ankle joints, this paper presents a task-oriented, customizable 2-DoF parallel elastic actuator. The proposed architecture achieves effective torque compensation across two degrees of freedom through the synergy of parallel gas springs and customized cam modules.

A comprehensive mathematical model was established and validated, extending the analysis from a single torque compensation unit to a coupled 2-DoF mechanism. By formulating and solving a constrained optimization problem, the optimal cam profiles were derived, providing a rigorous foundation for the proposed design workflow. The detailed hardware implementation, including component selection and CAD integration, has been finalized. Simulation results under various task-specific torque references demonstrate both the mechanical feasibility of the mechanism and the effectiveness of the optimization-based design framework.

Our future work will focus on the fabrication of a single-leg prototype based on the established CAD assembly. Extensive physical experiments will be conducted to quantify the actual torque compensation performance and the percentage of energy efficiency improvement across diverse locomotion tasks.

\section*{Acknowledgment}
The authors would like to express their sincere gratitude to Siddharth Deore and Weijie Wang for their assistance in solving the optimization problems and conducting the simulations. 

\bibliographystyle{./IEEEtran}
\bibliography{./IEEEabrv,./refs}

@INPROCEEDINGS{Eleg,
  author={Z. {Ren} and W. {Roozing} and N. G. {Tsagarakis}},
  booktitle={2018 IEEE-RAS 18th International Conference on Humanoid Robots (Humanoids)}, 
  title={The eLeg: A Novel Efficient Leg Prototype Powered by Adjustable Parallel Compliant Actuation Principles}, 
  year={2018},
  pages={1-9},
  doi={10.1109/IROS.2013.6696472},
  ISSN={2164-0580},
  month={Nov},}

@article{Pratt2004,
author = {Pratt, Jerry and Krupp, Benjamin},
year = {2004},
month = {09},
pages = {},
title = {Series Elastic Actuators for legged robots},
journal = {Proceedings of SPIE - The International Society for Optical Engineering},
doi = {10.1117/12.548000}
}

@article{Roozing2019,
author = {Roozing, Wesley and Ren, Zeyu and Tsagarakis, Nikos},
year = {2019},
month = {12},
pages = {},
title = {An Efficient Leg with Series-Parallel and Biarticular Compliant Actuation: Design Optimisation, Modelling, and Control of the eLeg},
journal = {The International Journal of Robotics Research},
doi = {10.1177/0278364919893762}
}

@ARTICLE{Seok6880316,

  author={Seok, Sangok and Wang, Albert and Chuah, Meng Yee and Hyun, Dong Jin and Lee, Jongwoo and Otten, David M. and Lang, Jeffrey H. and Kim, Sangbae},

  journal={IEEE/ASME Transactions on Mechatronics}, 

  title={Design Principles for Energy-Efficient Legged Locomotion and Implementation on the MIT Cheetah Robot}, 

  year={2015},

  volume={20},

  number={3},

  pages={1117-1129},

  doi={10.1109/TMECH.2014.2339013}}

@InProceedings{Xing101007,
author="Xing, Boyang
and Liu, Yufei
and Wang, Zhirui
and Liang, Zhenjie
and Zhao, Jianxin
and Su, Bo
and Jiang, Lei",
editor="Liu, Xin-Jun
and Nie, Zhenguo
and Yu, Jingjun
and Xie, Fugui
and Song, Rui",
title="The Moco-Minitaur: A Low-Cost Direct-Drive Quadruped Robot for Dynamic Locomotion",
booktitle="Intelligent Robotics and Applications",
year="2021",
publisher="Springer International Publishing",
address="Cham",
pages="378--389",
isbn="978-3-030-89095-7"
}

@article{Nan2021,
author = {Nan, Fang and Kolvenbach, Hendrik and Hutter, Marco},
year = {2021},
month = {12},
pages = {1-1},
title = {A Reconfigurable Leg for Walking Robots},
volume = {PP},
journal = {IEEE Robotics and Automation Letters},
doi = {10.1109/LRA.2021.3139379}
}

@ARTICLE{Wensing7827048,
  author={Wensing, Patrick M. and Wang, Albert and Seok, Sangok and Otten, David and Lang, Jeffrey and Kim, Sangbae},
  journal={IEEE Transactions on Robotics}, 
  title={Proprioceptive Actuator Design in the MIT Cheetah: Impact Mitigation and High-Bandwidth Physical Interaction for Dynamic Legged Robots}, 
  year={2017},
  volume={33},
  number={3},
  pages={509-522},
  doi={10.1109/TRO.2016.2640183}}

@article{CasADi2018,
author = {Andersson, Joel and Gillis, Joris and Horn, Greg and Rawlings, James and Diehl, Moritz},
year = {2018},
month = {07},
pages = {},
title = {CasADi: a software framework for nonlinear optimization and optimal control},
volume = {11},
journal = {Mathematical Programming Computation},
doi = {10.1007/s12532-018-0139-4}
}

@article{tsagarakis2017walk,
  title={Walk-man: A high-performance humanoid platform for realistic environments},
  author={Tsagarakis, Nikolaos G and Caldwell, Darwin G and Negrello, Francesca and Choi, Wooseok and Baccelliere, Lorenzo and Loc, Vo-Gia and Noorden, Jerryll and Muratore, Luca and Margan, Alessio and Cardellino, Alberto and others},
  journal={Journal of Field Robotics},
  volume={34},
  number={7},
  pages={1225--1259},
  year={2017},
  publisher={Wiley Online Library}
}

@inproceedings{malik2024intelligent,
  title={Intelligent humanoid robots in manufacturing},
  author={Malik, Ali Ahmad and Masood, Tariq and Brem, Alexander},
  booktitle={Companion of the 2024 ACM/IEEE International Conference on Human-Robot Interaction},
  pages={20--27},
  year={2024}
}

@article{collins2015reducing,
  title={Reducing the energy cost of human walking using an unpowered exoskeleton},
  author={Collins, Steven H and Wiggin, M Bruce and Sawicki, Gregory S},
  journal={Nature},
  volume={522},
  number={7555},
  pages={212--215},
  year={2015},
  publisher={Nature Publishing Group UK London}
}

@article{mathews2022design,
  title={Design of parallel variable stiffness actuators},
  author={Mathews, Chase W and Braun, David J},
  journal={IEEE Transactions on Robotics},
  volume={39},
  number={1},
  pages={768--782},
  year={2022},
  publisher={IEEE}
}

@article{Zhang2021Exo-Muscle:Knee,
    title = {{Exo-Muscle: A Semi-Rigid Assistive Device for the Knee}},
    year = {2021},
    journal = {IEEE Robotics and Automation Letters},
    author = {Zhang, Yifang and Ajoudani, Arash and Tsagarakis, Nikos G.},
    number = {4},
    month = {10},
    pages = {8514--8521},
    volume = {6},
    publisher = {Institute of Electrical and Electronics Engineers Inc.},
    doi = {10.1109/LRA.2021.3100609},
    issn = {23773766}
}

@article{wang2022design,
  title={Design and control of a series--parallel elastic actuator for a weight-bearing exoskeleton robot},
  author={Wang, Tianshuo and Zheng, Tianjiao and Zhao, Sikai and Sui, Dongbao and Zhao, Jie and Zhu, Yanhe},
  journal={Sensors},
  volume={22},
  number={3},
  pages={1055},
  year={2022},
  publisher={MDPI}
}

@article{zhang2024novel,
  title={A novel passive parallel elastic actuation principle for load compensation in legged robots},
  author={Zhang, Yifang and Jiang, Jingcheng and Tsagarakis, Nikos G},
  journal={IEEE Robotics and Automation Letters},
  volume={9},
  number={10},
  pages={8881--8888},
  year={2024},
  publisher={IEEE}
}

@inproceedings{fu2024energy,
  title={Energy Minimization using Custom-Designed Magnetic-Spring Actuators},
  author={Fu, Yue Yang and Kilic, Ali U and Braun, David J},
  booktitle={2024 IEEE/RSJ International Conference on Intelligent Robots and Systems (IROS)},
  pages={3534--3539},
  year={2024},
  organization={IEEE}
}

\end{document}